\documentclass[conference, a4paper]{IEEEtran}
\IEEEoverridecommandlockouts
\usepackage{cite}
\usepackage{amsmath,amssymb,amsfonts}
\usepackage{algorithmic}
\usepackage{graphicx}
\usepackage{textcomp}
\usepackage{xcolor}
\usepackage[hidelinks]{hyperref}
\usepackage[nolist]{acronym}
\usepackage{multirow}
\usepackage{array}
\usepackage{import}
\usepackage{tikz}
\usetikzlibrary{calc}
\usepackage{pgfplots}
\usetikzlibrary{plotmarks}
\pgfplotsset{compat=newest}
\usetikzlibrary{patterns}
\usetikzlibrary{calc, pgfplots.dateplot}
\usetikzlibrary{external}
\usetikzlibrary{quotes, angles}

\ifCLASSOPTIONcompsoc
\usepackage[caption=false,font=footnotesize,labelfont=sf,textfont=sf]{subfig}
\else
\usepackage[caption=false,font=footnotesize]{subfig}
\fi

\def\BibTeX{{\rm B\kern-.05em{\sc i\kern-.025em b}\kern-.08em
    T\kern-.1667em\lower.7ex\hbox{E}\kern-.125emX}}

\makeatletter
\newcommand{\linebreakand}{%
  \end{@IEEEauthorhalign}
  \hfill\mbox{}\par
  \mbox{}\hfill\begin{@IEEEauthorhalign}
}
\makeatother

\begin{document}

\newcolumntype{L}[1]{>{\raggedright\let\newline\\\arraybackslash\hspace{0pt}}m{#1}}
\newcolumntype{C}[1]{>{\centering\let\newline\\\arraybackslash\hspace{0pt}}m{#1}}
\newcolumntype{R}[1]{>{\raggedleft\let\newline\\\arraybackslash\hspace{0pt}}m{#1}}

\newcommand{\includetikz}[1]{
    \tikzsetnextfilename{./tikz/#1}
    \input{./tikz/#1.tikz}
}

\newcommand\paulo[1]{{\textcolor{orange}{Paulo: #1}}}
\newcommand\alceu[1]{{\textcolor{blue}{Alceu: #1}}}
\newcommand\eduardo[1]{{\textcolor{blue}{#1}}}

\begin{acronym}[TDMA]
    \acro{ASR-Norm}{Adaptive Standardization and Rescaling Normalization}
    \acro{k-NN}{k-Nearest Neighbors}
    \acro{i.i.d}{Independent and Identically Distributed}
\end{acronym}

\title{Distance Functions and Normalization Under Stream Scenarios}

% \author{\IEEEauthorblockN{Anonymous Authors}}

% \author{\IEEEauthorblockN{Eduardo V. L. Barboza}
% \IEEEauthorblockA{\textit{Department of Informatics} \\
% \textit{Universidade Federal do Paraná}\\
% Curitiba, Brazil \\
% eduardo.barboza@ufpr.br}
% \and
% \IEEEauthorblockN{Paulo R. Lisboa de Almeida}
% \IEEEauthorblockA{\textit{Department of Informatics} \\
% \textit{Universidade Federal do Paraná}\\
% Curitiba, Brazil \\
% paulorla@ufpr.br}
% \and
% \IEEEauthorblockN{Alceu de Souza Britto Jr.}
% \IEEEauthorblockA{\textit{Graduate Program in Informatics} \\
% \textit{Pontífica Universidade Católica do Paraná}\\
% Curitiba, Brazil \\
% \textit{Universidade Estadual de Ponta Grossa}\\
% Ponta Grossa, Brazil \\
% alceu@ppgia.pucpr.br}%\and
% \linebreakand
% \IEEEauthorblockN{Rafael M. O. Cruz}
% \IEEEauthorblockA{\textit{LIVIA} \\
% \textit{École de Technologie Supérieure}\\
% Quebec, Canada \\
% rafael.menelau-cruz@etsmtl.ca}
% }

\author{\IEEEauthorblockN{Eduardo V. L. Barboza\IEEEauthorrefmark{1}, Paulo R. Lisboa de Almeida\IEEEauthorrefmark{1}, Alceu de Souza Britto Jr.\IEEEauthorrefmark{2}\IEEEauthorrefmark{3} and Rafael M. O. Cruz\IEEEauthorrefmark{4}}
\IEEEauthorblockA{\IEEEauthorrefmark{1}Department of Informatics, Universidade Federal do Paraná, Curitiba (PR), Brazil\\
Email: \{eduardo.barboza, paulorla\}@ufpr.br}
\IEEEauthorblockA{\IEEEauthorrefmark{2}Graduate Program in Informatics, Pontífica Universidade Católica do Paraná, Curitiba (PR), Brazil}
\IEEEauthorblockA{\IEEEauthorrefmark{3}Universidade Estadual de Ponta Grossa, Ponta Grossa (PR), Brazil\\
Email: alceu@ppgia.puc.br}
\IEEEauthorblockA{\IEEEauthorrefmark{4}École de Technologie Supérieure, Université du Québec, Montréal (QC), Canada\\
Email: rafael.menelau-cruz@etsmtl.ca}}

\maketitle

\begin{abstract}
Data normalization is an essential task when modeling a classification system. When dealing with data streams, data normalization becomes especially challenging since we may not know in advance the properties of the features, such as their minimum/maximum values, and these properties may change over time. We compare the accuracies generated by eight well-known distance functions in data streams without normalization, normalized considering the statistics of the first batch of data received, and considering the previous batch received. We argue that experimental protocols for streams that consider the full stream as normalized are unrealistic and can lead to biased and poor results. Our results indicate that using the original data stream without applying normalization, and the Canberra distance, can be a good combination when no information about the data stream is known beforehand.
\end{abstract}


\section{Introduction}\label{sec:introduction}

When dealing with data streams, we face the scenario where new instances arrive over time. The stream size and the rate at which new instances arrive are usually unknown. Under such circumstances, classifiers are often updated over time since, at the beginning of the stream, only a few samples covering a small portion of the classification space are known.

Data normalization in such cases is a challenge if we do not have guarantees about the range of values generated for each feature. In other words, how can we apply normalization techniques, such as the min-max, in a possibly infinite stream without knowing beforehand the minimum/maximum values of each feature? 

It is essential to consider these points when dealing with classifiers that depend on data normalization or in the presence of concept drifts, where the range of the features (besides other properties) may change over time \cite{lu2018}. Some authors propose approaches to deal with streams that rely on normalizing the entire data stream or proceed to execute experimental protocols using normalized datasets -- e.g., most datasets currently available to test streams at the MOA website \cite{moa} are normalized. This may be unrealistic and lead to data leakage problems \cite{kaufmanEtAl2012}.

In this paper, we evaluate eight distance functions under different stream scenarios to give light on the following Research Questions:

\begin{itemize}
    \item \textbf{RQ1} -- Does the normalization policy influence the classifier's competence in data streams?
    \item \textbf{RQ2} -- Does the distance function matter when classifying data streams?
\end{itemize}

%In our tests, we show that the normalization of the full stream can sometimes lead to worse results. We also show that using the original data stream without normalization can lead to realistic and good results. We also show that distances such as the Cosine and Standardized Euclidean can lead to poor results. During the experiments, the Canberra distance led to the best results in the majority of the tests.

The answers to these questions are based on a robust experimental protocol composed of synthetic and real-world datasets. We confirm that the normalization of the entire stream can sometimes lead to worse results. The experiments have shown that when the classifier is retrained using the most recent data, using the original data stream without normalization combined with the Canberra distance function can provide more realistic and better results. Moreover, distances such as the Cosine and Standardized Euclidean can be more sensitive to feature changes over time than Manhattan and Canberra distances.

The remaining of this paper is structured into four sections. Section~\ref{sec:definitions} introduces the distance functions and the min-max normalization strategy evaluated in this paper. Section~\ref{sec:relatedWork} presents the related works. Section~\ref{sec:experiments} presents our experimental protocol, the test results, and a discussion about the observed results. Finally, Section~\ref{sec:conclusion} brings our conclusion and perspectives on future work.

\section{Definitions}\label{sec:definitions}

% In this Section, we present some concepts used in the remainder of the paper. In Section \ref{subsec:minmax}, we present the \textit{min-max} normalization that will be used throughout this work. In Section \ref{subsec:metrics}, we present eight distance functions that will be evaluated in this paper.

\subsection{Min-Max normalization}\label{subsec:minmax}

Throughout this paper, we employ the \textit{min-max} normalization, which is one of the most common normalization techniques, as it is simple to compute and to understand. The \textit{min-max} is defined as

\begin{equation}
    x_{ij} = \frac{x_{ij}-min_j}{max_j - min_j}
    \label{eq:minmax}
\end{equation}

\noindent where $j$ is the index of the $jth$ feature of the instance $x_{i}$. The $max_j$ and $min_j$ are the maximum and minimum values of the $jth$ features -- these values are often found by scanning the entire training set. A con of the min-max normalization is that it is sensitive to outliers, as it deals with minimum and maximum values.

\subsection{Distance Functions}\label{subsec:metrics}

We can define a distance function as a mathematical measure that quantifies how far apart two objects are \cite{cha2007}. Consider two instances $x = [x_1, x_2,\dots, x_n ]$ and $y = [y_1, y_2, \dots, y_n]$, where $x_i$ and $y_i$ is one of the $n$ features of $x$ or $y$, respectively. Finding a representative distance between $x$ and $y$ can be challenging if we consider that different features may lie in different ranges (non-normalized data), the presence of categorical data, missing points, computational cost, etc. 

In this paper, we consider only distance functions that deal with numerical data and assume that no missing features are present. Table \ref{tb:distanceMetrics} contains a list of the distance functions considered in this paper, where $d(x,y)$ is the distance between the instances $x$ and $y$.

{
\begin{table}[htpb]
\centering
\caption{Distance Functions.}
\renewcommand{\arraystretch}{3.0}
\begin{tabular}{ll}
Distance Function & Equation \\\hline
Euclidean & $\displaystyle d(x,y)=\sqrt{\sum_{i=1}^{n}(x_i-y_i)^2} $ \\
Manhattan & $\displaystyle d(x,y)=\sum_{i=1}^{n}|x_i-y_i|$ \\
Chebyshev & $\displaystyle d(x,y)=\max_{i=1}^n(|x_i-y_i|) $\\
Minkowski & $\displaystyle d(x,y)=(\sum^{n}_{i=1}|x_i-y_i|^p)^{\frac{1}{p}}$\\
Cosine & $\displaystyle d(x,y)=\frac{x \cdot y}{||x||\times||y||} $\\
Mahalanobis & $\displaystyle d(x,y)=\sqrt{(x-y)^T C^{-1} (x-y)}$\\
Standardized Euclidean & $\displaystyle d(x,y)=\sqrt{\sum_{i=1}^{n}\frac{(x_i-y_i)^2}{V}}$\\
Canberra & $\displaystyle d(x,y)=\sum^{n}_{i=1}\frac{|x_i-y_i|}{|x_i|+|y_i|}$ \\\hline
\end{tabular}

\label{tb:distanceMetrics}
\end{table}
}

\begin{figure}
    \begin{tikzpicture}

\begin{axis}[xlabel = $x$,
 ylabel = $y$,
  xmin=0,xmax=4,
  ymin=0,ymax=4,
 % ymajorgrids=true,
 % xmajorgrids=true,
 % grid style=dashed,
 label style={font=\tiny},
 tick label style={font=\tiny},
 legend style={font=\tiny},
 legend cell align=left,
 legend pos=south east]

  \draw[-,black] (1,3) -- (3,2);

  \addlegendimage{line width=0.3mm,color=black}
  \addlegendentry{Euclidean}
  
  \fill (1,3) circle[radius=2pt];
  \fill (3,2) circle[radius=2pt];

  \draw (1, 3.1) node[anchor=west] {$x$};
  \draw (3, 2) node[anchor=west] {$y$};

  \draw[-, blue, dotted] (1, 3) -- (1,2) -- (3, 2);
  \addlegendimage{line width=0.3mm,color=blue, style=dotted}
  \addlegendentry{Manhattan}

  \draw[-, cyan, dashdotted] (1, 3) -- (3,3);
  \addlegendimage{line width=0.3mm,color=cyan, style=dashdotted}
  \addlegendentry{Chebyshev}

  \draw[-, black, dotted]
    (0, 0) coordinate (a) node[right] {}
    -- (1,3) coordinate (b) node[left] {}
    (0,0) coordinate (c) node[above right] {} -- (3, 2) coordinate (d) node[right] {}
    pic["$\alpha$", draw=red, <->, style=solid, angle eccentricity=1.2, angle radius=1cm, "$\theta$"]
    {angle=d--c--b};

  \addlegendimage{line width=0.3mm,color=red, style=<->}
  \addlegendentry{Cosine}  
  
\end{axis}
    
\end{tikzpicture}
    \caption{Distance Functions in a 2-dimensional Space.}
    \label{fig:distances}
\end{figure}
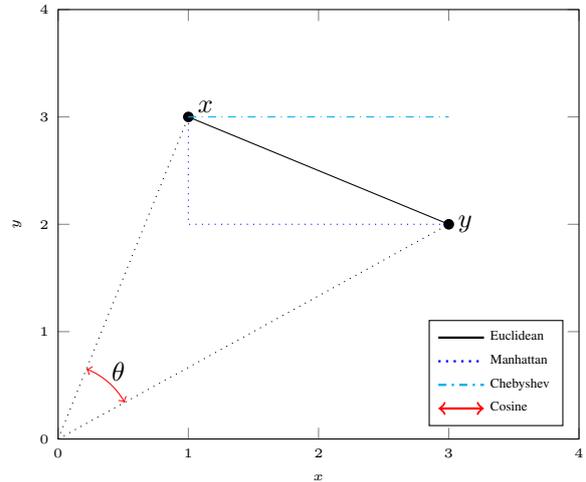

Figure \ref{fig:distances} shows an example of the distance functions for two points $x$ and $y$ in a 2-dimensional space. The Euclidean Distance is the most intuitive distance metric between two points, as it calculates a straight line between them. The Manhattan Distance, also known as taxicab geometry, considers a straight route between the points. It calculates the sum of the absolute differences between the features of $x$ and $y$. The Chebyshev Distance considers the maximum absolute difference between the features.

The Euclidean, Manhattan, and Chebyshev distance functions belong to the Minkowski family \cite{cha2007}, where when $p=1$ corresponds to the Manhattan Distance, and $p=2$ is the Euclidean Distance. When $p$ tends to infinity, it is similar to the Chebyshev Distance. Choosing the right value for the parameter $p$ in Minkowski Distance also might influence the performance \cite{Rodrigues_2018, lu2016}. We use $p=1.5$ in our experiments since it is a middle term between Euclidean and Manhattan distances.

The Cosine distance takes into account the angle between data points instead of the distance between them. The Mahalanobis Distance uses a covariance matrix $C$ when calculating the distance -- i.e., it takes into account the relation between the features. If the Matrix $C$ in Mahalanobis distance is the identity matrix, it is like features have no relation with each other, and the Mahalanobis distance is similar to the Euclidean Distance. Standardized Euclidean is the same as the Euclidean Distance, but it divides the difference in features by the variance $V$ of data. Finally, the Canberra distance divides the absolute difference of features by the sum of absolute features.

\section{Related Work}\label{sec:relatedWork}

The authors in \cite{AlmeidaEtAl2020} show some problems related to unrealistic scenarios when modeling streams. The authors demonstrate that commonly used benchmarks in state-of-the-art datasets contain a high serial dependence. Thus experimental protocols that rely on the \ac{i.i.d} assumption may lead to biased results.

When dealing with data streams, authors often use some technique to normalize the data stream using a fixed-size window. The window is moved when new data is available, and the data is normalized considering the statistics of the current window \cite{ogasawaraEtAl2010,guptaHewett2019}.

The authors in \cite{ogasawaraEtAl2010} use disjoint sliding windows to estimate the global min-max values for normalization. They show that their adaptive normalization technique got better results than other normalization methods such as min-max normalization, z-score, decimal scaling, and min-max with a sliding window in the tested datasets. To save computational resources, the authors in \cite{guptaHewett2019} propose a technique to update the normalization only when the statistics in the current and previous windows are above a specified threshold. They compared their proposed method using the min-max scaler using different approaches for windowing in data normalization. They analyzed a base policy where the data range is known for the whole dataset and compared the error between different policies. Their method got the least root mean squared error to this base policy.

In \cite{xinjie2021}, authors proposed \ac{ASR-Norm}, an adaptive normalization method where statistics for standardization and rescaling are learned through neural networks. It outperforms Batch Normalization, Instance Normalization, and Switchable Normalization.

Methods that utilize distance functions to deal with data streams, like in \cite{losing2016, ALMEIDA2018}, may be impacted by how we deal with data normalization. Authors in \cite{amorim2023} tested some Machine Learning algorithms with five different scaling techniques and affirmed that the chosen scaling technique influences the performance and the best one changes with the dataset used. 
% Authors got good results with the Manhattan distance.

In \cite{Nayak_2022}, the authors assessed the performance of different distance functions using a \ac{k-NN} for classifying stars. In their experiment, the Cosine distance function got the best accuracy with $k=9$. Authors in \cite{Batista2009} analyzed three different distance functions for Distance-Weighted \ac{k-NN}: Heterogeneous Euclidean-Overlap metric, Heterogeneous Euclidean-VDM metric, and Heterogeneous Manhattan-Overlap metric The authors did not find a significant difference between them.

Authors in \cite{yean2018} tested different $k$ values and distance functions for \ac{k-NN} on classifying emotional electroencephalogram between stroke and normal people, and got the best results with the Manhattan distance. They also concluded that the distance functions have different performances depending on the situation. In \cite{mladenova2021}, different normalization techniques, distance functions, and \ac{k-NN} configurations were analyzed to classify fake news. The combination of Robust Scaler, Chebyshev distance, and $k=34$ got the best result. 

Many works have their datasets already normalized \cite{mehmood2021, krawczyk2018}. The point is that many of these datasets may be normalized using the whole stream, e.g., the datasets available in the MOA repository \cite{moa}, and this is not realistic. Authors in \cite{xinjie2021} argue that most works regarding normalization do not study the capacity of generalization in non-stationary environments. Care must be taken with normalization, as shown by \cite{singh2020}, who argues that normalization sometimes leads to worse performance -- a conclusion that we get to in this work as well. This proves that the assumption that normalization improves performance does not hold in all cases.

In this work, we chose the min-max normalization technique, applied different policies for normalizing data, and tested how \ac{k-NN} behaves with different distance functions in different scenarios inside data streams. To the best of our knowledge, there is no work studying the impact of different normalization policies and distance functions under data streams.

\section{Experiments}\label{sec:experiments}

% In this Section, we present our experimental protocol and the results regarding synthetic and real-world data. Our objective is to compare different distance functions and normalization policies under stream scenarios.

\subsection{Experimental Protocol}\label{sec:experimental_protocol}

In this study, we run some experiments to evaluate the impact of different distance functions in different scenarios. We do it by comparing the accuracies of a 3NN (\ac{k-NN} with $k=3$) with the distance functions described in Section \ref{subsec:metrics}. We chose the 3NN classifier since it is a weak learner that directly depends on the distance functions to classify the instances. The rationale is to perceive accuracy changes better when using different distance functions.

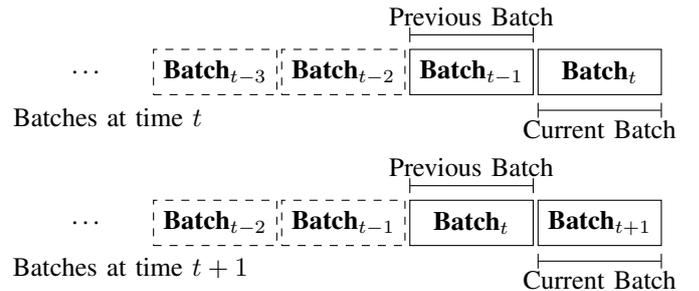
\begin{figure}[htpb]
    \centering
	\begin{tikzpicture}
        \matrix [nodes=draw,column sep=0.5mm, minimum height=0.6cm, minimum width={width("Batch$t-1$")+2pt}, anchor=west] at (0,0) {
        \node[draw=none]{\dots}; &
        \node[dashed]{\textbf{Batch}$_{t-3}$}; &
        \node[dashed]{\textbf{Batch}$_{t-2}$}; &
        \node(tp1){\textbf{Batch}$_{t-1}$}; &
        \node(t){\textbf{Batch}$_{t}$}; &
        %\node[dashed](tn1){\textbf{Batch}$_{t+1}$};
        \\
        };

        \draw [|-|,black] ($(tp1.south west) - (0,-0.8cm)$) -- ($(tp1.south east) - (0,-0.8cm)$) node [above,midway] {Previous Batch};
        
        \draw [|-|,black] ($(t.south west) - (0,0.2cm)$) -- ($(t.south east) - (0,0.2cm)$) node [below,midway] {Current Batch};

        %\draw [|-|,black] ($(tn1.south west) - (0,-0.8cm)$) -- ($(tn1.south east) - (0,-0.8cm)$) node [above,midway] {Next Batch};

        \node[draw=none] at (1.2, -0.6cm){Batches at time $t$};

        \matrix [nodes=draw,column sep=0.5mm, minimum height=0.6cm, minimum width={width("Batch$t-1$")+2pt}, anchor=west] at (0,-2cm){
            \node[draw=none]{\dots}; &
            \node[dashed]{\textbf{Batch}$_{t-2}$}; & \node(t2p1)[dashed]{\textbf{Batch}$_{t-1}$}; & \node(t2){\textbf{Batch}$_{t}$}; &
            \node(t2n1){\textbf{Batch}$_{t+1}$}; \\
        };

        \draw [|-|,black] ($(t2.south west) - (0,-0.8cm)$) -- ($(t2.south east) - (0,-0.8cm)$) node [above,midway] {Previous Batch};

        \draw [|-|,black] ($(t2n1.south west) - (0,0.2cm)$) -- ($(t2n1.south east) - (0,0.2cm)$) node [below,midway] {Current Batch};

        \node[draw=none] at (1.5, -2.6cm){Batches at time $t+1$};
        
    \end{tikzpicture}\\
	\caption{Scheme of a Stream of Batches.}
	\label{fig:batchStream}
\end{figure}

During the tests, we split the data into batches containing 1,000 samples. When a new batch is given at time $t+1$, the true labels of the instances of the previous batch, received at $t$, are given. The task of the classifier is to predict the instances available in the most recent (current) batch received. Figure \ref{fig:batchStream} shows a scheme of a stream of batches.

We test the impact of normalization for the different distance metrics under four scenarios:

\begin{enumerate}
    \item The \textit{original} stream, without any normalization.
    \item The statistics of the \textit{first batch} are used to normalize the remaining batches using the min-max approach.
    \item The statistics of the \textit{previous} batch received are used to normalize the \textit{current} one.
    \item The \textit{full} stream is normalized using the min-max approach.
\end{enumerate}

Notice that the normalization of the entire stream (item 4) is often unfeasible in the real world, as there is often no way to know a stream's minimum and maximum values. We use this approach to compare with items 1-3 and to demonstrate how the results may be biased under such an unrealistic scenario.
%For the tests, we compare a classifier trained only with the first batch, which we call \textit{First Train}, and a classifier that is \textit{retrained} with the previous batch received. 
When we used the approach for normalizing data in the previous batch, the scaler was retrained before updating the model.

Table \ref{tb:datasets} contains a summary of the datasets employed in this work. The datasets are available in well-known repositories, such as the UCI \cite{uci}, the MOA \cite{moa}, and the OpenML \cite{OpenML} repositories. When a missing feature is present, it was replaced by the mean value of the whole dataset -- again, this is not possible in the real world, it was made here for testing purposes. We use the SEA Concepts \cite{street2001} as an artificial dataset in the tests reported in Section \ref{sec:changing_range}. This dataset contains three randomly generated real features $f1$, $f2$, and $f3 \in [0, 10]$. We consider only the first concept, where the generated sample belongs to the positive class if $f1 + f2 \leq 8$, or to the negative class otherwise (the feature $f3$ is noise).

\begin{table}[ht]
\caption{Datasets.}
\setlength{\tabcolsep}{4pt}
\begin{tabular}{l r r r r}\hline
Dataset & \multicolumn{1}{c}{\# Feat.} & \multicolumn{1}{c}{\# Instances} & \multicolumn{1}{c}{\# Classes} & Available at \\
\hline
SEA Concepts     & 3           & 40,000        & 2          &     See \cite{street2001}  \\
Electricity      & 8           & 45,312        & 2          & MOA Repository   \\
Airlines         & 8           & 539,383       & 2          & OpenML   \\
Pokerhand        & 10          & 1,025,010     & 10         & UCI Repository       \\
Forest Covertype & 54          & 581,012       & 7          & UCI Repository     \\
Gas Sensor       & 128         & 13,910        & 6          & UCI Repository      \\\hline
\end{tabular}

\label{tb:datasets}
\end{table}

The reported results are an average of 30 trials.

%For the experiments, we use a machine with an Intel Core i7 with 16GiB of DRAM and Ubuntu 22.04 as the operating System. The reported results are an average of 30 trials.
\subsection{Tests using Synthetic Data}\label{sec:changing_range}

In this Section, we evaluate the accuracy of the classifier in an environment where the range of the features changes over time. It may occur in the real world due to, for example, variations in temperature with the change of seasons or to a faulty sensor \cite{demvsarBosnic2018}. In each run, 40,000 instances of the SEA dataset are generated.

We first made a test varying the $f1$ feature. From the 1st to the 10,000th instance, no modification is made. From the 10,001st to the 20,000th instance, the instances are multiplied by 10, from the 20,001st to the 30,000th by 100, and from the 30,001st to the 40,000th by 1,000. We follow the same protocol in a test where the $f3$ feature is modified over time -- notice that differently from $f1$, the feature $f3$ is non-informative.

The results regarding a classifier that is trained with the first batch of the stream and never updated are available in Table \ref{tab:results_sea_first_train}. When we varied the range of feature $f1$, we can observe that no distance metric performs well without normalization. The results improve significantly when the normalization is made, considering the \textit{previous} batch.

When we vary the non-informative $f3$ feature, the normalization that considers the previous batch leads to better results. Interestingly, the normalization of the full stream led to worse results when using the Chebyshev, Canberra, and Standardized Euclidean distances. The normalization of the full stream led to better results in some scenarios, such as when using the Euclidean distance. Thus, besides being unrealistic, the normalization of the entire stream may lead to biased results for better or worse, depending on the distance function.

\begin{table}[htpb]
\centering
\caption{Accuracies on SEA Dataset (Trained in First Batch). The Minkowski was computed using $p=1.5$.}
\setlength{\tabcolsep}{1pt} % Default value: 6pt
\begin{tabular}{l|c c c c|c c c c}
\hline
& \multicolumn{4}{c|}{Varying $f1$} & \multicolumn{4}{c}{Varying $f3$} \\
& Origin. & First B. & Prev. & Full & Origin. & First B. & Prev. & Full \\\hline

Euclidean & 0.756 & 0.755 & \textbf{0.939} & \textbf{0.752} & 0.906 & 0.819 & 0.954 & 0.976 \\
Manhattan &  0.756 & 0.755 & 0.938 & 0.722 & 0.963 & 0.964 & \textbf{0.958} & 0.976 \\
Cosine & 0.607 & 0.617 & 0.748 & 0.617 & 0.442 & 0.441 & 0.735 & 0.350 \\
Chebyshev & 0.756 & 0.755 & \textbf{0.939} & 0.693 & 0.714 & 0.715 & 0.933 & 0.746 \\
Mahalanobis & 0.756 & 0.755 & \textbf{0.939} & 0.722 & 0.745 & 0.760 & 0.950 & \textbf{0.977} \\
Std. Eucl.& 0.756 & 0.755 & \textbf{0.939} & 0.688 & 0.806 & 0.820 & 0.954 & 0.582 \\
Minkowski & 0.756 & 0.755 & \textbf{0.939} & 0.728 & 0.906 & 0.909 & 0.956 & \textbf{0.977} \\
Canberra & \textbf{0.762} & \textbf{0.761} & 0.922 & 0.714 & \textbf{0.967} & \textbf{0.967} & 0.941 & 0.965 \\
Average & 0.738 & 0.735 & 0.913 & 0.704 & 0.806 & 0.799 & 0.923 & 0.819 \\\hline

\end{tabular}

\label{tab:results_sea_first_train}
\end{table}

When we consider a classifier retrained with the previous batch, shown in Table \ref{tab:results_sea_retrained}, we can reach similar results with the normalization that considers the previous batch being the best one. It is also possible to notice that the Canberra distance reached the best results when the original dataset (without any normalization) was used. The Manhattan distance is the second best in such a scenario. This result corroborates with \cite{amorim2023}, where the Manhattan distance worked well when no normalization was made. In all tests, the Cosine distance showed the worst results.

\begin{table}[htpb]
\centering
\caption{Accuracies on SEA Dataset (Retrained). The Minkowski was computed using $p=1.5$.}
\setlength{\tabcolsep}{1pt} % Default value: 6pt
\begin{tabular}{l|c c c c|c c c c}
\hline
& \multicolumn{4}{c|}{Varying $f1$} & \multicolumn{4}{c}{Varying $f3$}\\

& Origin. & First B. & Prev. & Full & Origin. & First B. & Prev. & Full \\\hline

Euclidean & 0.866 & 0.866 & 0.939 & 0.811 & 0.825 & 0.825 & 0.954 & 0.972 \\
Manhattan & 0.875 & 0.875 & 0.938 & 0.819 & 0.842 & 0.844 & \textbf{0.957} & 0.971 \\
Cosine & 0.751 & 0.753 & 0.749 & 0.734 & 0.736 & 0.736 & 0.736 & 0.707 \\
Chebyshev & 0.858 & 0.858 & \textbf{0.940} & 0.807 & 0.814 & 0.815 & 0.935 & 0.968 \\
Mahalanobis & 0.866 & 0.866 & 0.939 & 0.722 & 0.824 & 0.824 & 0.951 & \textbf{0.977} \\
Std. Eucl.& 0.866 & 0.866 & 0.939 & 0.859 & 0.825 & 0.825 & 0.954 & 0.825 \\
Minkowski & 0.869 & 0.869 & 0.939 & 0.813 & 0.830 & 0.831 & 0.956 & 0.972 \\
Canberra & \textbf{0.923} & \textbf{0.922} & 0.923 & \textbf{0.923} & \textbf{0.942} & \textbf{0.941} & 0.941 & 0.942 \\
Average & 0.859 & 0.859 & 0.913 & 0.811 & 0.829 & 0.830 & 0.923 & 0.917 \\\hline

\end{tabular}

\label{tab:results_sea_retrained}
\end{table}

In Figure \ref{fig:accuracies_varying_f1_SEA_no_var_concept}, we show the accuracy reached in each batch of data for the tested distance functions when the $f1$ feature is changed. In all scenarios, the cosine distance led to the worst results, even before the change in the feature range. We can observe that (apart from the cosine distance), all distance functions led to similar results when the classifier is not retrained (Figures \ref{subfig:seaNoNoiseFTNoNormVarF1} and \ref{subfig:seaNoNoiseFTFBNormVarF1}).

When the model is retrained, but no normalization is made (Figure \ref{subfig:seaNoNoiseRTNoNormVarF1}), the Canberra distance leads to the best results, followed by the Manhattan distance. When both the classifier and the normalization are retrained using the previous batch (Figure \ref{subfig:seaNoNoiseRTPBNormVarF1}), once again, all distance functions except the cosine seem to behave similarly.

\begin{figure*}[htpb]
 	\centering
 	{
\definecolor{mycolor1}{rgb}{0.92900,0.69400,0.12500}%
\begin{tikzpicture}

\begin{axis}[%
hide axis,
xmin=0,
xmax=1,
ymin=0,
ymax=1,
ylabel={$f_2$},
axis x line*=bottom,
axis y line*=left,
legend style={legend cell align=left,align=left,draw=black,legend columns=8, column sep=2.5pt, font=\footnotesize} 
]

\addlegendimage{color=black, solid, mark=square, mark options={solid}}
\addlegendentry{euclidean};

\addlegendimage{color=blue, dotted, mark=o,mark options={solid}}
\addlegendentry{manhattan};

\addlegendimage{color=cyan, dashdotted, mark=x,mark options={solid}}
\addlegendentry{chebyshev};

\addlegendimage{color=orange, solid, mark=triangle,mark options={solid}}
\addlegendentry{mahalanobis};

\addlegendimage{color=magenta, dotted, mark=diamond,mark options={solid}}
\addlegendentry{std. eucl.};

\addlegendimage{color=gray, solid, mark=Mercedes star, mark options={solid}}
\addlegendentry{minkowski};

\addlegendimage{color=black, solid, mark=otimes, mark options={solid}}
\addlegendentry{canberra};

\addlegendimage{color=red, dashed, mark=x,mark options={solid}}
\addlegendentry{cosine};

\end{axis}
\end{tikzpicture}%
 	}\\
    {
 		\subfloat[First Train without normalization]{
            \label{subfig:seaNoNoiseFTNoNormVarF1}
 			
    \tikzsetnextfilename{./tikz/seaNoDriftFTNoNormVarF1}
    \begin{tikzpicture}

\begin{axis}[%
width=6.0cm,
height=3.0cm,
at={(0,0)},
scale only axis,
xmin=0,
xlabel={Batch},
ymin=0.2,
ymax=1.0,
ylabel={Accuracy},
legend style={at={(0.03,0.57)}, anchor=south west, font=\footnotesize},
y label style={at={(axis description cs:-0.14,.5)},anchor=south},
]
%\addplot [color=black,solid,mark=o,mark options={solid},mark repeat=24, mark phase=0]

\draw [dashed] (10, 0) -- (10,1);
\draw [dashed] (20, 0) -- (20,1);
\draw [dashed] (30, 0) -- (30,1);

\addplot [color=black, solid, each nth point=1,mark=square,mark options={solid},mark repeat=10,  mark phase=0] table [x=idx, y=euclidean, col sep=comma] {tikz/accuracies_ft_varying_f1_nonorm.csv};

\addplot [color=blue, dotted, each nth point=1, mark=o,mark options={solid},mark repeat=5,  mark phase=2] table [x=idx, y=manhattan, col sep=comma] {tikz/accuracies_ft_varying_f1_nonorm.csv};

\addplot [color=cyan, dashdotted, each nth point=1, mark=x,mark options={solid},mark repeat=5,  mark phase=4] table [x=idx, y=chebyshev, col sep=comma] {tikz/accuracies_ft_varying_f1_nonorm.csv};

\addplot [color=orange, solid, each nth point=1, mark=triangle,mark options={solid},mark repeat=5,  mark phase=6] table [x=idx, y=mahalanobis, col sep=comma] {tikz/accuracies_ft_varying_f1_nonorm.csv};

\addplot [color=magenta, dotted, each nth point=1, mark=diamond,mark options={solid},mark repeat=5,  mark phase=8] table [x=idx, y=seuclidean, col sep=comma] {tikz/accuracies_ft_varying_f1_nonorm.csv};

\addplot [color=gray, solid, each nth point=1, mark=Mercedes star, mark options={solid},mark repeat=5, mark phase=10] table [x=idx, y=minkowski, col sep=comma] {tikz/accuracies_ft_varying_f1_nonorm.csv};

\addplot [color=black, solid, each nth point=1,  mark=otimes, mark options={solid},mark repeat=5, mark phase=12] table [x=idx, y=canberra, col sep=comma] {tikz/accuracies_ft_varying_f1_nonorm.csv};

\addplot [color=red, dashed, each nth point=1, mark=x,mark options={solid},mark repeat=5,  mark phase=0] table [x=idx, y=cosine, col sep=comma] {tikz/accuracies_ft_varying_f1_nonorm.csv};

\end{axis}
\end{tikzpicture}

 		}
 	}
     {
 		\subfloat[First Train Normalized First Batch]{
            \label{subfig:seaNoNoiseFTFBNormVarF1}
 			
    \tikzsetnextfilename{./tikz/seaNoDriftFTFBNormVarF1}
    \begin{tikzpicture}

\begin{axis}[%
width=6.0cm,
height=3.0cm,
at={(0,0)},
scale only axis,
xmin=0,
xlabel={Batch},
ymin=0.2,
ymax=1.0,
ylabel={Accuracy},
legend style={at={(0.03,0.57)}, anchor=south west, font=\footnotesize},
y label style={at={(axis description cs:-0.14,.5)},anchor=south},
]
%\addplot [color=black,solid,mark=o,mark options={solid},mark repeat=24, mark phase=0]

\draw [dashed] (10, 0) -- (10,1);
\draw [dashed] (20, 0) -- (20,1);
\draw [dashed] (30, 0) -- (30,1);

\addplot [color=black, solid, each nth point=1,mark=square,mark options={solid},mark repeat=10,  mark phase=0] table [x=idx, y=euclidean, col sep=comma] {tikz/accuracies_ft_varying_f1_fbnorm.csv};

\addplot [color=blue, dotted, each nth point=1, mark=o,mark options={solid},mark repeat=5,  mark phase=2] table [x=idx, y=manhattan, col sep=comma] {tikz/accuracies_ft_varying_f1_fbnorm.csv};

\addplot [color=cyan, dashdotted, each nth point=1, mark=x,mark options={solid},mark repeat=5,  mark phase=4] table [x=idx, y=chebyshev, col sep=comma] {tikz/accuracies_ft_varying_f1_fbnorm.csv};

\addplot [color=orange, solid, each nth point=1, mark=triangle,mark options={solid},mark repeat=5,  mark phase=6] table [x=idx, y=mahalanobis, col sep=comma] {tikz/accuracies_ft_varying_f1_fbnorm.csv};

\addplot [color=magenta, dotted, each nth point=1, mark=diamond,mark options={solid},mark repeat=5,  mark phase=8] table [x=idx, y=seuclidean, col sep=comma] {tikz/accuracies_ft_varying_f1_fbnorm.csv};

\addplot [color=gray, solid, each nth point=1, mark=Mercedes star, mark options={solid},mark repeat=5, mark phase=10] table [x=idx, y=minkowski, col sep=comma] {tikz/accuracies_ft_varying_f1_fbnorm.csv};

\addplot [color=black, solid, each nth point=1,  mark=otimes, mark options={solid},mark repeat=5, mark phase=12] table [x=idx, y=canberra, col sep=comma] {tikz/accuracies_ft_varying_f1_fbnorm.csv};

\addplot [color=red, dashed, each nth point=1, mark=x,mark options={solid},mark repeat=5,  mark phase=0] table [x=idx, y=cosine, col sep=comma] {tikz/accuracies_ft_varying_f1_fbnorm.csv};

\end{axis}
\end{tikzpicture}

 		}
 	}
 	{
 		\subfloat[Retrained without normalization]{
            \label{subfig:seaNoNoiseRTNoNormVarF1}
 			
    \tikzsetnextfilename{./tikz/seaNoDriftRTNoNormVarF1}
    \begin{tikzpicture}

\begin{axis}[%
width=6.0cm,
height=3.0cm,
at={(0,0)},
scale only axis,
xmin=0,
xlabel={Batch},
ymin=0.2,
ymax=1.0,
ylabel={Accuracy},
legend style={at={(0.03,0.57)}, anchor=south west, font=\footnotesize},
y label style={at={(axis description cs:-0.14,.5)},anchor=south},
]
%\addplot [color=black,solid,mark=o,mark options={solid},mark repeat=24, mark phase=0]

\draw [dashed] (10, 0) -- (10,1);
\draw [dashed] (20, 0) -- (20,1);
\draw [dashed] (30, 0) -- (30,1);

\addplot [color=black, solid, each nth point=1,mark=square,mark options={solid},mark repeat=10,  mark phase=0] table [x=idx, y=euclidean, col sep=comma] {tikz/accuracies_rt_varying_f1_nonorm.csv};

\addplot [color=blue, dotted, each nth point=1, mark=o,mark options={solid},mark repeat=5,  mark phase=2] table [x=idx, y=manhattan, col sep=comma] {tikz/accuracies_rt_varying_f1_nonorm.csv};

\addplot [color=cyan, dashdotted, each nth point=1, mark=x,mark options={solid},mark repeat=5,  mark phase=4] table [x=idx, y=chebyshev, col sep=comma] {tikz/accuracies_rt_varying_f1_nonorm.csv};

\addplot [color=orange, solid, each nth point=1, mark=triangle,mark options={solid},mark repeat=5,  mark phase=6] table [x=idx, y=mahalanobis, col sep=comma] {tikz/accuracies_rt_varying_f1_nonorm.csv};

\addplot [color=magenta, dotted, each nth point=1, mark=diamond,mark options={solid},mark repeat=5,  mark phase=8] table [x=idx, y=seuclidean, col sep=comma] {tikz/accuracies_rt_varying_f1_nonorm.csv};

\addplot [color=gray, solid, each nth point=1, mark=Mercedes star, mark options={solid},mark repeat=5, mark phase=10] table [x=idx, y=minkowski, col sep=comma] {tikz/accuracies_rt_varying_f1_nonorm.csv};

\addplot [color=black, solid, each nth point=1,  mark=otimes, mark options={solid},mark repeat=5, mark phase=12] table [x=idx, y=canberra, col sep=comma] {tikz/accuracies_rt_varying_f1_nonorm.csv};

\addplot [color=red, dashed, each nth point=1, mark=x,mark options={solid},mark repeat=5,  mark phase=0] table [x=idx, y=cosine, col sep=comma] {tikz/accuracies_rt_varying_f1_nonorm.csv};

\end{axis}
\end{tikzpicture}

 		}
 	}
     {
 		\subfloat[Retrained Normalized Previous Batch]{
            \label{subfig:seaNoNoiseRTPBNormVarF1}
 			
    \tikzsetnextfilename{./tikz/seaNoDriftRTPBNormVarF1}
    \begin{tikzpicture}

\begin{axis}[%
width=6.0cm,
height=3.0cm,
at={(0,0)},
scale only axis,
xmin=0,
xlabel={Batch},
ymin=0.2,
ymax=1.0,
ylabel={Accuracy},
legend style={at={(0.03,0.57)}, anchor=south west, font=\footnotesize},
y label style={at={(axis description cs:-0.14,.5)},anchor=south},
]
%\addplot [color=black,solid,mark=o,mark options={solid},mark repeat=24, mark phase=0]

\draw [dashed] (10, 0) -- (10,1);
\draw [dashed] (20, 0) -- (20,1);
\draw [dashed] (30, 0) -- (30,1);

\addplot [color=black, solid, each nth point=1,mark=square,mark options={solid},mark repeat=10,  mark phase=0] table [x=idx, y=euclidean, col sep=comma] {tikz/accuracies_rt_varying_f1_pbnorm.csv};

\addplot [color=blue, dotted, each nth point=1, mark=o,mark options={solid},mark repeat=5,  mark phase=2] table [x=idx, y=manhattan, col sep=comma] {tikz/accuracies_rt_varying_f1_pbnorm.csv};

\addplot [color=cyan, dashdotted, each nth point=1, mark=x,mark options={solid},mark repeat=5,  mark phase=4] table [x=idx, y=chebyshev, col sep=comma] {tikz/accuracies_rt_varying_f1_pbnorm.csv};

\addplot [color=orange, solid, each nth point=1, mark=triangle,mark options={solid},mark repeat=5,  mark phase=6] table [x=idx, y=mahalanobis, col sep=comma] {tikz/accuracies_rt_varying_f1_pbnorm.csv};

\addplot [color=magenta, dotted, each nth point=1, mark=diamond,mark options={solid},mark repeat=5,  mark phase=8] table [x=idx, y=seuclidean, col sep=comma] {tikz/accuracies_rt_varying_f1_pbnorm.csv};

\addplot [color=gray, solid, each nth point=1, mark=Mercedes star, mark options={solid},mark repeat=5, mark phase=10] table [x=idx, y=minkowski, col sep=comma] {tikz/accuracies_rt_varying_f1_pbnorm.csv};

\addplot [color=black, solid, each nth point=1,  mark=otimes, mark options={solid},mark repeat=5, mark phase=12] table [x=idx, y=canberra, col sep=comma] {tikz/accuracies_rt_varying_f1_pbnorm.csv};

\addplot [color=red, dashed, each nth point=1, mark=x,mark options={solid},mark repeat=5,  mark phase=0] table [x=idx, y=cosine, col sep=comma] {tikz/accuracies_rt_varying_f1_pbnorm.csv};

\end{axis}
\end{tikzpicture}

 		}
 	}
 	\caption{Accuracies in SEA Varying Range of $f1$.}
 	\label{fig:accuracies_varying_f1_SEA_no_var_concept}
 \end{figure*}

When the range of a non-significant feature ($f3$) varies, we can see different behavior between the distance functions in Figure \ref{fig:accuracies_varying_f3_SEA_no_var_concept}. Interestingly, the Canberra and Manhattan distance functions did not show any accuracy drop even in the moments when the range is changed -- for instance, in the 10th batch (except for the scenario retrained without normalization, where the Manhattan distance presents accuracy drops).

 \begin{figure*}[htpb]
 	\centering
 	{
\definecolor{mycolor1}{rgb}{0.92900,0.69400,0.12500}%
\begin{tikzpicture}

\begin{axis}[%
hide axis,
xmin=0,
xmax=1,
ymin=0,
ymax=1,
ylabel={$f_2$},
axis x line*=bottom,
axis y line*=left,
legend style={legend cell align=left,align=left,draw=black,legend columns=8, column sep=2.5pt, font=\footnotesize} 
]

\addlegendimage{color=black, solid, mark=square, mark options={solid}}
\addlegendentry{euclidean};

\addlegendimage{color=blue, dotted, mark=o,mark options={solid}}
\addlegendentry{manhattan};

\addlegendimage{color=cyan, dashdotted, mark=x,mark options={solid}}
\addlegendentry{chebyshev};

\addlegendimage{color=orange, solid, mark=triangle,mark options={solid}}
\addlegendentry{mahalanobis};

\addlegendimage{color=magenta, dotted, mark=diamond,mark options={solid}}
\addlegendentry{std. eucl.};

\addlegendimage{color=gray, solid, mark=Mercedes star, mark options={solid}}
\addlegendentry{minkowski};

\addlegendimage{color=black, solid, mark=otimes, mark options={solid}}
\addlegendentry{canberra};

\addlegendimage{color=red, dashed, mark=x,mark options={solid}}
\addlegendentry{cosine};

\end{axis}
\end{tikzpicture}%
 	}\\
    {
 		\subfloat[First Train without normalization]{\label{subfig:seaNoNoiseFTNoNorm}
 			
    \tikzsetnextfilename{./tikz/seaNoDriftFTNoNormVarF3}
    \begin{tikzpicture}

\begin{axis}[%
width=6.0cm,
height=3.0cm,
at={(0,0)},
scale only axis,
xmin=0,
xlabel={Batch},
ymin=0.2,
ymax=1.0,
ylabel={Accuracy},
legend style={at={(0.03,0.57)}, anchor=south west, font=\footnotesize},
y label style={at={(axis description cs:-0.14,.5)},anchor=south},
]
%\addplot [color=black,solid,mark=o,mark options={solid},mark repeat=24, mark phase=0]

\draw [dashed] (10, 0) -- (10,1);
\draw [dashed] (20, 0) -- (20,1);
\draw [dashed] (30, 0) -- (30,1);

\addplot [color=black, solid, each nth point=1,mark=square,mark options={solid},mark repeat=10,  mark phase=0] table [x=idx, y=euclidean, col sep=comma] {tikz/accuracies_ft_varying_f3_nonorm.csv};

\addplot [color=blue, dotted, each nth point=1, mark=o,mark options={solid},mark repeat=5,  mark phase=2] table [x=idx, y=manhattan, col sep=comma] {tikz/accuracies_ft_varying_f3_nonorm.csv};

\addplot [color=cyan, dashdotted, each nth point=1, mark=x,mark options={solid},mark repeat=5,  mark phase=4] table [x=idx, y=chebyshev, col sep=comma] {tikz/accuracies_ft_varying_f3_nonorm.csv};

\addplot [color=orange, solid, each nth point=1, mark=triangle,mark options={solid},mark repeat=5,  mark phase=6] table [x=idx, y=mahalanobis, col sep=comma] {tikz/accuracies_ft_varying_f3_nonorm.csv};

\addplot [color=magenta, dotted, each nth point=1, mark=diamond,mark options={solid},mark repeat=5,  mark phase=8] table [x=idx, y=seuclidean, col sep=comma] {tikz/accuracies_ft_varying_f3_nonorm.csv};

\addplot [color=gray, solid, each nth point=1, mark=Mercedes star, mark options={solid},mark repeat=5, mark phase=10] table [x=idx, y=minkowski, col sep=comma] {tikz/accuracies_ft_varying_f3_nonorm.csv};

\addplot [color=black, solid, each nth point=1,  mark=otimes, mark options={solid},mark repeat=5, mark phase=12] table [x=idx, y=canberra, col sep=comma] {tikz/accuracies_ft_varying_f3_nonorm.csv};

\addplot [color=red, dashed, each nth point=1, mark=x,mark options={solid},mark repeat=5,  mark phase=0] table [x=idx, y=cosine, col sep=comma] {tikz/accuracies_ft_varying_f3_nonorm.csv};

\end{axis}
\end{tikzpicture}

 		}
 	}
     {
 		\subfloat[First Train Normalized First Batch]{\label{subfig:seaNoNoiseFTNoNorm2}
 			
    \tikzsetnextfilename{./tikz/seaNoDriftFTFBNormVarF3}
    \begin{tikzpicture}

\begin{axis}[%
width=6.0cm,
height=3.0cm,
at={(0,0)},
scale only axis,
xmin=0,
xlabel={Batch},
ymin=0.2,
ymax=1.0,
ylabel={Accuracy},
legend style={at={(0.03,0.57)}, anchor=south west, font=\footnotesize},
y label style={at={(axis description cs:-0.14,.5)},anchor=south},
]
%\addplot [color=black,solid,mark=o,mark options={solid},mark repeat=24, mark phase=0]

\draw [dashed] (10, 0) -- (10,1);
\draw [dashed] (20, 0) -- (20,1);
\draw [dashed] (30, 0) -- (30,1);

\addplot [color=black, solid, each nth point=1,mark=square,mark options={solid},mark repeat=10,  mark phase=0] table [x=idx, y=euclidean, col sep=comma] {tikz/accuracies_ft_varying_f3_fbnorm.csv};

\addplot [color=blue, dotted, each nth point=1, mark=o,mark options={solid},mark repeat=5,  mark phase=2] table [x=idx, y=manhattan, col sep=comma] {tikz/accuracies_ft_varying_f3_fbnorm.csv};

\addplot [color=cyan, dashdotted, each nth point=1, mark=x,mark options={solid},mark repeat=5,  mark phase=4] table [x=idx, y=chebyshev, col sep=comma] {tikz/accuracies_ft_varying_f3_fbnorm.csv};

\addplot [color=orange, solid, each nth point=1, mark=triangle,mark options={solid},mark repeat=5,  mark phase=6] table [x=idx, y=mahalanobis, col sep=comma] {tikz/accuracies_ft_varying_f3_fbnorm.csv};

\addplot [color=magenta, dotted, each nth point=1, mark=diamond,mark options={solid},mark repeat=5,  mark phase=8] table [x=idx, y=seuclidean, col sep=comma] {tikz/accuracies_ft_varying_f3_fbnorm.csv};

\addplot [color=gray, solid, each nth point=1, mark=Mercedes star, mark options={solid},mark repeat=5, mark phase=10] table [x=idx, y=minkowski, col sep=comma] {tikz/accuracies_ft_varying_f3_fbnorm.csv};

\addplot [color=black, solid, each nth point=1,  mark=otimes, mark options={solid},mark repeat=5, mark phase=12] table [x=idx, y=canberra, col sep=comma] {tikz/accuracies_ft_varying_f3_fbnorm.csv};

\addplot [color=red, dashed, each nth point=1, mark=x,mark options={solid},mark repeat=5,  mark phase=0] table [x=idx, y=cosine, col sep=comma] {tikz/accuracies_ft_varying_f3_fbnorm.csv};

\end{axis}
\end{tikzpicture}

 		}
 	}
 	{
 		\subfloat[Retrained without normalization]{\label{subfig:seaNoNoiseFTNoNorm3}
 			
    \tikzsetnextfilename{./tikz/seaNoDriftRTNoNormVarF3}
    \begin{tikzpicture}

\begin{axis}[%
width=6.0cm,
height=3.0cm,
at={(0,0)},
scale only axis,
xmin=0,
xlabel={Batch},
ymin=0.2,
ymax=1.0,
ylabel={Accuracy},
legend style={at={(0.03,0.57)}, anchor=south west, font=\footnotesize},
y label style={at={(axis description cs:-0.14,.5)},anchor=south},
]
%\addplot [color=black,solid,mark=o,mark options={solid},mark repeat=24, mark phase=0]

\draw [dashed] (10, 0) -- (10,1);
\draw [dashed] (20, 0) -- (20,1);
\draw [dashed] (30, 0) -- (30,1);

\addplot [color=black, solid, each nth point=1,mark=square,mark options={solid},mark repeat=10,  mark phase=0] table [x=idx, y=euclidean, col sep=comma] {tikz/accuracies_rt_varying_f3_nonorm.csv};

\addplot [color=blue, dotted, each nth point=1, mark=o,mark options={solid},mark repeat=5,  mark phase=2] table [x=idx, y=manhattan, col sep=comma] {tikz/accuracies_rt_varying_f3_nonorm.csv};

\addplot [color=cyan, dashdotted, each nth point=1, mark=x,mark options={solid},mark repeat=5,  mark phase=4] table [x=idx, y=chebyshev, col sep=comma] {tikz/accuracies_rt_varying_f3_nonorm.csv};

\addplot [color=orange, solid, each nth point=1, mark=triangle,mark options={solid},mark repeat=5,  mark phase=6] table [x=idx, y=mahalanobis, col sep=comma] {tikz/accuracies_rt_varying_f3_nonorm.csv};

\addplot [color=magenta, dotted, each nth point=1, mark=diamond,mark options={solid},mark repeat=5,  mark phase=8] table [x=idx, y=seuclidean, col sep=comma] {tikz/accuracies_rt_varying_f3_nonorm.csv};

\addplot [color=gray, solid, each nth point=1, mark=Mercedes star, mark options={solid},mark repeat=5, mark phase=10] table [x=idx, y=minkowski, col sep=comma] {tikz/accuracies_rt_varying_f3_nonorm.csv};

\addplot [color=black, solid, each nth point=1,  mark=otimes, mark options={solid},mark repeat=5, mark phase=12] table [x=idx, y=canberra, col sep=comma] {tikz/accuracies_rt_varying_f3_nonorm.csv};

\addplot [color=red, dashed, each nth point=1, mark=x,mark options={solid},mark repeat=5,  mark phase=0] table [x=idx, y=cosine, col sep=comma] {tikz/accuracies_rt_varying_f3_nonorm.csv};

\end{axis}
\end{tikzpicture}

 		}
 	}
     {
 		\subfloat[Retrained Normalized Previous Batch]{\label{subfig:seaNoNoiseFTNoNorm4}
 			
    \tikzsetnextfilename{./tikz/seaNoDriftRTPBNormVarF3}
    \begin{tikzpicture}

\begin{axis}[%
width=6.0cm,
height=3.0cm,
at={(0,0)},
scale only axis,
xmin=0,
xlabel={Batch},
ymin=0.2,
ymax=1.0,
ylabel={Accuracy},
legend style={at={(0.03,0.57)}, anchor=south west, font=\footnotesize},
y label style={at={(axis description cs:-0.14,.5)},anchor=south},
]
%\addplot [color=black,solid,mark=o,mark options={solid},mark repeat=24, mark phase=0]

\draw [dashed] (10, 0) -- (10,1);
\draw [dashed] (20, 0) -- (20,1);
\draw [dashed] (30, 0) -- (30,1);

\addplot [color=black, solid, each nth point=1,mark=square,mark options={solid},mark repeat=10,  mark phase=0] table [x=idx, y=euclidean, col sep=comma] {tikz/accuracies_rt_varying_f3_pbnorm.csv};

\addplot [color=blue, dotted, each nth point=1, mark=o,mark options={solid},mark repeat=5,  mark phase=2] table [x=idx, y=manhattan, col sep=comma] {tikz/accuracies_rt_varying_f3_pbnorm.csv};

\addplot [color=cyan, dashdotted, each nth point=1, mark=x,mark options={solid},mark repeat=5,  mark phase=4] table [x=idx, y=chebyshev, col sep=comma] {tikz/accuracies_rt_varying_f3_pbnorm.csv};

\addplot [color=orange, solid, each nth point=1, mark=triangle,mark options={solid},mark repeat=5,  mark phase=6] table [x=idx, y=mahalanobis, col sep=comma] {tikz/accuracies_rt_varying_f3_pbnorm.csv};

\addplot [color=magenta, dotted, each nth point=1, mark=diamond,mark options={solid},mark repeat=5,  mark phase=8] table [x=idx, y=seuclidean, col sep=comma] {tikz/accuracies_rt_varying_f3_pbnorm.csv};

\addplot [color=gray, solid, each nth point=1, mark=Mercedes star, mark options={solid},mark repeat=5, mark phase=10] table [x=idx, y=minkowski, col sep=comma] {tikz/accuracies_rt_varying_f3_pbnorm.csv};

\addplot [color=black, solid, each nth point=1,  mark=otimes, mark options={solid},mark repeat=5, mark phase=12] table [x=idx, y=canberra, col sep=comma] {tikz/accuracies_rt_varying_f3_pbnorm.csv};

\addplot [color=red, dashed, each nth point=1, mark=x,mark options={solid},mark repeat=5,  mark phase=0] table [x=idx, y=cosine, col sep=comma] {tikz/accuracies_rt_varying_f3_pbnorm.csv};

\end{axis}
\end{tikzpicture}

 		}
 	}
 	\caption{Accuracies in SEA Varying Range of $f3$.}
 	\label{fig:accuracies_varying_f3_SEA_no_var_concept}
 \end{figure*}
\subsection{Tests using Real-World Datasets} \label{sec:real_experiments}

In this section, we check the behavior of the distance functions in five real-world datasets. Results are displayed in Table \ref{tab:results_real}. In these experiments, the classifiers are always retrained using the previous batch. First, considering the average results, we can observe that the (unrealistic) normalization of the full datasets led to better results in the Airlines and Gas Sensor datasets, and it worsened the results in the remaining datasets.
Using the original dataset without normalization often led to better results when considering the normalization techniques.
 In Table \ref{tab:victories_norm}, we show a count of the number of wins each normalization technique led.

{
\begin{table*}[htpb]
\centering
\caption{Accuracies on Real Datasets. The classifier is retrained using the previous batch.}
\setlength{\tabcolsep}{1pt} % Default value: 6pt
\begin{tabular}{l|c c c c|c c c c|c c c c|c c c c|c c c c}
\hline
 & \multicolumn{4}{c|}{\textbf{Electricity}} & \multicolumn{4}{c|}{\textbf{Airlines}} & \multicolumn{4}{c|}{\textbf{Pokerhand}} & \multicolumn{4}{c|}{\textbf{Forest Covertype}} & \multicolumn{4}{c}{\textbf{Gas Sensor}}\\
& Origin. & First B. & Prev. & Full & Origin. & First B. & Prev. & Full & Origin. & First B. & Prev. & Full & Origin. & First B. & Prev. & Full & Origin. & First B. & Prev. & Full\\\hline

Euclidean & 0.739 & 0.719 & 0.680 & 0.682 & 0.576 & 0.597 & 0.580 & 0.604 & 0.504 & 0.481 & 0.481 & 0.481 & 0.918 & 0.897 & 0.897 & 0.897 & 0.579 & 0.604 & 0.539 & 0.622 \\
Manhattan & 0.741 & 0.719 & 0.681 & 0.693 & 0.582 & \textbf{0.612} & 0.595 & \textbf{0.614} & 0.502 & 0.482 & 0.482 & 0.482 & \textbf{0.919} & \textbf{0.900} & \textbf{0.901} & \textbf{0.900} & 0.586 & 0.621 & 0.542 & \textbf{0.633} \\
Cosine & 0.721 & 0.700 & 0.658 & 0.679 & 0.574 & 0.598 & 0.579 & 0.599 & 0.501 & 0.479 & 0.479 & 0.479 & 0.913 & 0.896 & 0.896 & 0.897 & \textbf{0.638} & 0.624 & 0.543 & 0.631 \\
Chebyshev & 0.732 & 0.715 & 0.669 & 0.671 & 0.573 & 0.581 & 0.572 & 0.595 & 0.504 & \textbf{0.484} & \textbf{0.484} & \textbf{0.484} & 0.914 & 0.886 & 0.883 & 0.887 & 0.575 & 0.559 & 0.485 & 0.596 \\
Mahalanobis & 0.739 & 0.699 & 0.669 & 0.666 & 0.559 & 0.605 & 0.582 & 0.607 & \textbf{0.509} & 0.479 & 0.479 & 0.479 & 0.892 & 0.889 & 0.891 & 0.885 & 0.493 & 0.513 & 0.466 & 0.537 \\
Std. Eucl. & 0.641 & 0.641 & 0.555 & 0.745 & 0.541 & 0.541 & 0.540 & 0.563 & 0.483 & 0.483 & 0.483 & 0.483 & 0.477 & 0.477 & 0.477 & 0.477 & 0.595 & 0.595 & 0.501 & 0.595 \\
Minkowski & 0.739 & 0.719 & 0.681 & 0.686 & 0.578 & 0.603 & 0.586 & 0.607 & 0.503 & 0.481 & 0.481 & 0.481 & \textbf{0.919} & 0.898 & 0.898 & 0.897 & 0.584 & 0.609 & \textbf{0.547} & 0.623  \\
Canberra & \textbf{0.742} & \textbf{0.744} & \textbf{0.687} & \textbf{0.749} & \textbf{0.603} & 0.602 & \textbf{0.598} & 0.603 & 0.489 & 0.483 & 0.483 & 0.483 & 0.894 & \textbf{0.900} & 0.898 & 0.899 & 0.631 & \textbf{0.626} & 0.539 & 0.630 \\
Average & 0.724 & 0.707 & 0.66 & 0.696 & 0.573 & 0.592 & 0.579 & 0.599 & 0.499 & 0.482 & 0.482 & 0.482 & 0.856 & 0.843 & 0.843 & 0.842 & 0.585 & 0.594 & 0.520 & 0.608 \\\hline

\end{tabular}

\label{tab:results_real}
\end{table*}
}

\begin{table}[htpb]
    \centering
    \caption{Number of Victories of Normalization Approaches (Full not included).}
    \begin{tabular}{l|c}\hline
        Normalization & Victories \\\hline
        Original & 3 \\
        First Batch & 2 \\
        Previous Batch & 0 \\\hline
    \end{tabular}
    
    \label{tab:victories_norm}
\end{table}

Note in Table \ref{tab:victories_norm} that, differently from the tests in Section \ref{sec:changing_range}, the normalization using the Previous Batch did not lead to the best results in any dataset. We hypothesize that this may have happened due to two possible factors: 1) The datasets tested do not present significative changes in the ranges of the features. 2) The datasets tested may present frequent changes (drifts) over time; thus, when the data is normalized with the previous batch, the ranges have already changed.

\begin{figure*}[htpb]
    \centering

    % \begin{align}
        \subfloat[Covertype.]{
        \label{subfig:box_covtype}
        \includegraphics[width=0.3\linewidth]{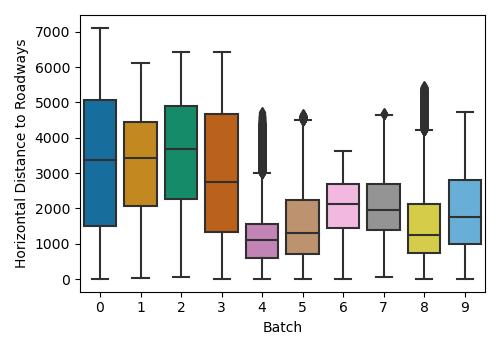}
        }
        \subfloat[Gas Sensor.]{
        \label{subfig:box_gassensor}
        \includegraphics[width=0.3\linewidth]{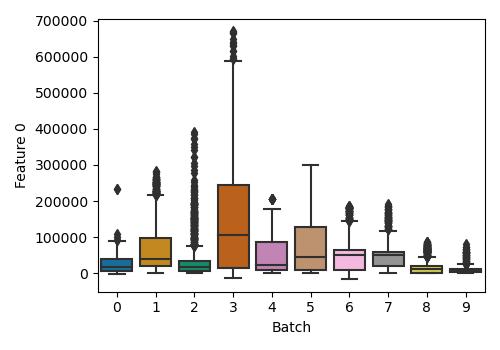}
        }        
        \subfloat[Electricity.]{
        \label{subfig:box_elec}
        \includegraphics[width=0.3\linewidth]{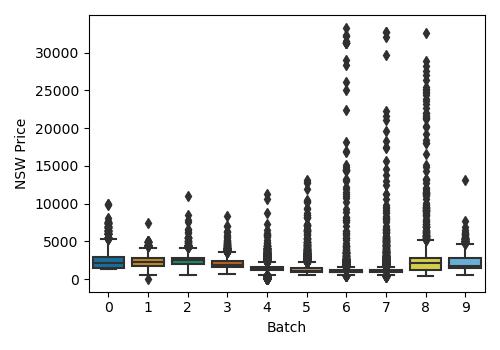}}
    % \end{align}
    
    \caption{Box Plots for Features in Datasets.}
    \label{fig:boxplots}
\end{figure*}

To better understand the results above, let us analyze the box plots of the features with the highest standard deviation of some datasets, divided into ten batches. In Figure \ref{subfig:box_covtype}, we show the box plot for one feature of the Forest Covertype dataset. We can see a higher variation after the 4th batch, where outliers start to arise. Notice that the min-max scaler is not optimal under the presence of outliers \cite{amorim2023}, which may explain its poor performance when compared with the original data. Similar analysis can be done for the Gas Sensor and Electricity datasets in Figures \ref{subfig:box_gassensor} and \ref{subfig:box_elec}, respectively.
Even though we are analyzing only one feature for each dataset in the plots of Figure \ref{fig:boxplots}, these insights suggest that analyzing these statistics for the remaining features, especially the highly correlated with the target class, can be a prospect of future work.

In Table \ref{tab:victories_dist}, we show how many times each distance metric led to the best result (we did not consider the results when taking the full normalization of the dataset). As one can observe, the Canberra distance, followed by the Manhattan distance, were the distances that led to the best results more often. In Table \ref{tab:mean_distances} we show the average results achieved by each distance metric in each dataset. The average considers the original dataset and the normalization that takes the first and previous batch (full is not considered here). 
% The Canberra and Manhattan distances often reached similar results.

\begin{table}[htpb]
    \centering
    \caption{Number of Victories of Each Distance Function.}
    \begin{tabular}{l|c}\hline
        Distance Function & Victories \\\hline
        Euclidean & 0 \\
        Manhattan & 4 \\
        Cosine & 1 \\
        Chebyshev & 2 \\
        Mahalanobis & 1 \\
        Std. Eucl. & 0 \\
        Minkowski & 2 \\
        Canberra & 7 \\\hline
    \end{tabular}
    
    \label{tab:victories_dist}
\end{table}

%It is worth noticing the poor performance of the Standardized Euclidean on the Forest Covertype dataset. This may be due to the fact that $44$ out $54$ features are binary, which may difficult the computation of the Variance $V$.

\begin{table}[htpb]
\centering
\caption{Accuracy Averages of Each Distance Function on Datasets (Full Normalization Not Included).}
\setlength{\tabcolsep}{1pt} % Default value: 6pt
\begin{tabular}{l|ccccc}\hline
                 & Electricity & Airlines & Pokerhand & Covertype & Gas Sensor\\\hline
Euclidean    & 0.713 & 0.584  & \textbf{0.489} & 0.904  & 0.574 \\
Manhattan    & 0.714 & 0.596 & \textbf{0.489} & 0.904 & 0.583 \\
Cosine       & 0.693 & 0.584 & 0.486 & 0.902 & \textbf{0.602} \\
Chebyshev    & 0.705 & 0.575 & 0.491 & 0.894 & 0.539 \\
Mahalanobis  & 0.702 & 0.582 & \textbf{0.489} & 0.891 & 0.491 \\
Std. Eucl.   & 0.612 & 0.541 & 0.483 & 0.477 & 0.564 \\
Minkowski    & 0.713 & 0.589 & 0.488 & \textbf{0.905} & 0.580 \\
Canberra     & \textbf{0.724} & \textbf{0.601} & 0.485 & 0.897 & 0.599 \\\hline
\end{tabular}

\label{tab:mean_distances}
\end{table}

\subsection{Discussion and Limitations}\label{subsec:discussion}

The results presented in Sections \ref{sec:changing_range} and \ref{sec:real_experiments} lead us to some interesting findings:

1) Besides being unrealistic, the normalization of the full dataset may lead to biased results. Surprisingly, the bias can affect the results negatively in some scenarios. This finding reinforces that experiments for data streams should avoid the full stream's normalization to create scenarios closer to the real world.

2) The normalization made using the information from the previous batch may be beneficial under scenarios where the features range changes severely over time (see Section \ref{sec:changing_range}). Nevertheless, when this may not happen, the original data may lead to better results. Thus, as a conservative approach, using the original data (without normalization) can be a good approach since it can lead to good results without the overhead of the data normalization.

3) The Canberra distance was the most resilient one, leading to good results in various scenarios. The Canberra distance is also simple to compute; thus, we indicate this as a default metric to be used under stream scenarios.

4) The cosine and Standardized Euclidean distances did not show good results for most tested scenarios. Thus, except for specific cases, these metrics should be avoided under stream scenarios.

With these findings, we can answer our research questions:

\textbf{RQ1 -- Does the normalization policy influence the classifier's competence in data streams?} Answer: yes, normalization policy can lead to different results under data streams. We suggest the usage of the original data since it got the best result in three out of five tests using real-world data. It is important to update the classifier with the most recent data, as shown in the results of Section \ref{sec:changing_range}.

\textbf{RQ2 -- Does the distance function matter when classifying data streams?} Answer: yes, and we suggest the usage of the Canberra distance. As shown in Table \ref{tab:victories_dist}, the Canberra distance got the most victories among the distance functions tested in this study -- 7 out of 15 tests.

It is important to mention some limitations of this work. First, we did not consider problems regarding a stream where the ranges of features change with a high frequency -- for instance, scenarios where the range may change for every batch. Although we have not carried out tests for such scenarios, the results of this work make us hypothesize that normalizing the stream using the previous batch may deteriorate the results. Second, this work does not cover streams of instances (instead of batches), where the overhead of updating the normalization over time may be prohibitive -- nevertheless, using the original data without normalization, as suggested in this work, may be a good start.

\section{Conclusion}\label{sec:conclusion}
    
In this study, we analyzed the impact of eight distance functions and three normalization approaches in data streams -- using the original data, the data normalized using the first batch, and the data normalized using the previous batch. Tests using one synthetic and five well-known real datasets showed us some interesting results.

First, we demonstrate that the normalization of the full dataset, besides being unrealistic, may lead to biased results. Surprisingly, the results can even be biased for the worse. We showed that different normalization policies may lead to different results under streams. Nevertheless, it is difficult to conclude the best normalization policy for a general case. We suggest that the usage of the original data without normalization can be a good conservative approach.

We also show that the Canberra distance function showed the best results in most of our tests, and thus we indicate this as a distance metric to be used when it is not possible to check in advance the properties of the stream. We also show that the Manhattan distance can lead to good results, and distances such as the cosine and Standardized Euclidean may lead to poor results in some streams.

In future works, we intend to test other scaling techniques, such as the z-score, as well as testing other classifiers and classification techniques that directly depend on the distance metrics. We also intend to test scenarios containing a stream of instances instead of batches and check the accuracy and overhead caused by the different distance metrics and normalization policies.

\section*{Acknowledgement} 

The authors would like to thank Conselho Nacional de Desenvolvimento Científico e Tecnológico (CNPq, Brazil, grant 405511/2022-1).

\bibliographystyle{IEEEtran}
\bibliography{referencias}

\begin{thebibliography}{10}
\providecommand{\url}[1]{#1}
\csname url@samestyle\endcsname
\providecommand{\newblock}{\relax}
\providecommand{\bibinfo}[2]{#2}
\providecommand{\BIBentrySTDinterwordspacing}{\spaceskip=0pt\relax}
\providecommand{\BIBentryALTinterwordstretchfactor}{4}
\providecommand{\BIBentryALTinterwordspacing}{\spaceskip=\fontdimen2\font plus
\BIBentryALTinterwordstretchfactor\fontdimen3\font minus
  \fontdimen4\font\relax}
\providecommand{\BIBforeignlanguage}[2]{{%
\expandafter\ifx\csname l@#1\endcsname\relax
\typeout{** WARNING: IEEEtran.bst: No hyphenation pattern has been}%
\typeout{** loaded for the language `#1'. Using the pattern for}%
\typeout{** the default language instead.}%
\else
\language=\csname l@#1\endcsname
\fi
#2}}
\providecommand{\BIBdecl}{\relax}
\BIBdecl

\bibitem{lu2018}
J.~Lu, A.~Liu, F.~Dong, F.~Gu, J.~Gama, and G.~Zhang, ``Learning under concept
  drift: A review,'' \emph{{IEEE} Transactions on Knowledge and Data
  Engineering}, pp. 1--1, 2018.

\bibitem{moa}
\BIBentryALTinterwordspacing
A.~Bifet, G.~Holmes, R.~Kirkby, and B.~Pfahringer, ``{MOA:} massive online
  analysis,'' \emph{J. Mach. Learn. Res.}, vol.~11, pp. 1601--1604, 2010.
  [Online]. Available: \url{https://dl.acm.org/doi/10.5555/1756006.1859903}
\BIBentrySTDinterwordspacing

\bibitem{kaufmanEtAl2012}
S.~Kaufman, S.~Rosset, C.~Perlich, and O.~Stitelman, ``Leakage in data mining:
  Formulation, detection, and avoidance,'' \emph{ACM Transactions on Knowledge
  Discovery from Data (TKDD)}, vol.~6, no.~4, pp. 1--21, 2012.

\bibitem{cha2007}
S.-H. Cha, ``Comprehensive survey on distance/similarity measures between
  probability density functions,'' \emph{Int. J. Math. Model. Meth. Appl.
  Sci.}, vol.~1, 01 2007.

\bibitem{Rodrigues_2018}
\BIBentryALTinterwordspacing
{\'{E}}.~Rodrigues, ``Combining minkowski and chebyshev: New distance proposal
  and survey of distance metrics using k-nearest neighbours classifier,''
  \emph{Pattern Recognition Letters}, vol. 110, pp. 66--71, jul 2018. [Online].
  Available: \url{https://doi.org/10.1016\%2Fj.patrec.2018.03.021}
\BIBentrySTDinterwordspacing

\bibitem{lu2016}
\BIBentryALTinterwordspacing
B.~Lu, M.~Charlton, C.~Brunsdon, and P.~Harris, ``The minkowski approach for
  choosing the distance metric in geographically weighted regression,''
  \emph{Int. J. Geogr. Inf. Sci.}, vol.~30, no.~2, p. 351–368, feb 2016.
  [Online]. Available: \url{https://doi.org/10.1080/13658816.2015.1087001}
\BIBentrySTDinterwordspacing

\bibitem{AlmeidaEtAl2020}
P.~R. Lisboa~de Almeida, L.~S. Oliveira, A.~d. Souza~Britto, and
  J.~Paul~Barddal, ``Naïve approaches to deal with concept drifts,'' in
  \emph{2020 IEEE International Conference on Systems, Man, and Cybernetics
  (SMC)}, 2020, pp. 1052--1059.

\bibitem{ogasawaraEtAl2010}
E.~Ogasawara, L.~C. Martinez, D.~De~Oliveira, G.~Zimbr{\~a}o, G.~L. Pappa, and
  M.~Mattoso, ``Adaptive normalization: A novel data normalization approach for
  non-stationary time series,'' in \emph{The 2010 International Joint
  Conference on Neural Networks (IJCNN)}.\hskip 1em plus 0.5em minus
  0.4em\relax IEEE, 2010, pp. 1--8.

\bibitem{guptaHewett2019}
V.~Gupta and R.~Hewett, ``Adaptive normalization in streaming data,'' in
  \emph{Proceedings of the 2019 3rd International Conference on Big Data
  Research}, 2019, pp. 12--17.

\bibitem{xinjie2021}
\BIBentryALTinterwordspacing
X.~Fan, Q.~Wang, J.~Ke, F.~Yang, B.~Gong, and M.~Zhou, ``Adversarially adaptive
  normalization for single domain generalization,'' 2021. [Online]. Available:
  \url{https://arxiv.org/abs/2106.01899}
\BIBentrySTDinterwordspacing

\bibitem{losing2016}
V.~Losing, B.~Hammer, and H.~Wersing, ``Knn classifier with self adjusting
  memory for heterogeneous concept drift,'' in \emph{2016 IEEE 16th
  International Conference on Data Mining (ICDM)}, 2016, pp. 291--300.

\bibitem{ALMEIDA2018}
\BIBentryALTinterwordspacing
P.~R. Almeida, L.~S. Oliveira, A.~S. Britto, and R.~Sabourin, ``Adapting
  dynamic classifier selection for concept drift,'' \emph{Expert Systems with
  Applications}, vol. 104, pp. 67--85, 2018. [Online]. Available:
  \url{https://www.sciencedirect.com/science/article/pii/S0957417418301611}
\BIBentrySTDinterwordspacing

\bibitem{amorim2023}
\BIBentryALTinterwordspacing
L.~B. {de Amorim}, G.~D. Cavalcanti, and R.~M. Cruz, ``The choice of scaling
  technique matters for classification performance,'' \emph{Applied Soft
  Computing}, vol. 133, p. 109924, 2023. [Online]. Available:
  \url{https://www.sciencedirect.com/science/article/pii/S1568494622009735}
\BIBentrySTDinterwordspacing

\bibitem{Nayak_2022}
\BIBentryALTinterwordspacing
S.~Nayak, M.~Bhat, N.~V.~S. Reddy, and B.~A. Rao, ``Study of distance metrics
  on k - nearest neighbor algorithm for star categorization,'' \emph{Journal of
  Physics: Conference Series}, vol. 2161, no.~1, p. 012004, jan 2022. [Online].
  Available: \url{https://dx.doi.org/10.1088/1742-6596/2161/1/012004}
\BIBentrySTDinterwordspacing

\bibitem{Batista2009}
G.~E. A. P.~A. Batista and D.~F. Silva, ``How k-nearest neighbor parameters
  affect its performance,'' in \emph{Argentine Symposium on Artificial
  Intelligence}, 2009.

\bibitem{yean2018}
C.~W. Yean, W.~Khairunizam, M.~I. Omar, M.~Murugappan, B.~S. Zheng, S.~A.
  Bakar, Z.~M. Razlan, and Z.~Ibrahim, ``Analysis of the distance metrics of
  knn classifier for eeg signal in stroke patients,'' in \emph{2018
  International Conference on Computational Approach in Smart Systems Design
  and Applications (ICASSDA)}, 2018, pp. 1--4.

\bibitem{mladenova2021}
T.~Mladenova and I.~Valova, ``Analysis of the knn classifier distance metrics
  for bulgarian fake news detection,'' in \emph{2021 3rd International Congress
  on Human-Computer Interaction, Optimization and Robotic Applications (HORA)},
  2021, pp. 1--4.

\bibitem{mehmood2021}
\BIBentryALTinterwordspacing
H.~Mehmood, P.~Kostakos, M.~Cortes, T.~Anagnostopoulos, S.~Pirttikangas, and
  E.~Gilman, ``Concept drift adaptation techniques in distributed environment
  for real-world data streams,'' \emph{Smart Cities}, vol.~4, no.~1, pp.
  349--371, 2021. [Online]. Available:
  \url{https://www.mdpi.com/2624-6511/4/1/21}
\BIBentrySTDinterwordspacing

\bibitem{krawczyk2018}
B.~Krawczyk, B.~Pfahringer, and M.~Woźniak, ``Combining active learning with
  concept drift detection for data stream mining,'' in \emph{2018 IEEE
  International Conference on Big Data (Big Data)}, 2018, pp. 2239--2244.

\bibitem{singh2020}
\BIBentryALTinterwordspacing
D.~Singh and B.~Singh, ``Investigating the impact of data normalization on
  classification performance,'' \emph{Applied Soft Computing}, vol.~97, p.
  105524, 2020. [Online]. Available:
  \url{https://www.sciencedirect.com/science/article/pii/S1568494619302947}
\BIBentrySTDinterwordspacing

\bibitem{uci}
\BIBentryALTinterwordspacing
D.~Dua and C.~Graff, ``{UCI} machine learning repository,'' 2017. [Online].
  Available: \url{http://archive.ics.uci.edu/ml}
\BIBentrySTDinterwordspacing

\bibitem{OpenML}
\BIBentryALTinterwordspacing
J.~Vanschoren, J.~N. van Rijn, B.~Bischl, and L.~Torgo, ``Openml: Networked
  science in machine learning,'' \emph{SIGKDD Explorations}, vol.~15, no.~2,
  pp. 49--60, 2013. [Online]. Available:
  \url{http://doi.acm.org/10.1145/2641190.2641198}
\BIBentrySTDinterwordspacing

\bibitem{street2001}
\BIBentryALTinterwordspacing
W.~N. Street and Y.~Kim, ``A streaming ensemble algorithm (sea) for large-scale
  classification,'' in \emph{Proceedings of the Seventh ACM SIGKDD
  International Conference on Knowledge Discovery and Data Mining}, ser. KDD
  '01.\hskip 1em plus 0.5em minus 0.4em\relax New York, NY, USA: Association
  for Computing Machinery, 2001, p. 377–382. [Online]. Available:
  \url{https://doi.org/10.1145/502512.502568}
\BIBentrySTDinterwordspacing

\bibitem{demvsarBosnic2018}
J.~Dem{\v{s}}ar and Z.~Bosni{\'c}, ``Detecting concept drift in data streams
  using model explanation,'' \emph{Expert Systems with Applications}, vol.~92,
  pp. 546--559, 2018.

\end{thebibliography}

\end{document}